\pdfoutput=1
\documentclass[11pt]{article}

\usepackage{authblk}
\usepackage[final]{acl}
\usepackage{multirow} 
\usepackage{times}
\usepackage{latexsym}
\usepackage{lipsum}

\usepackage[utf8]{inputenc}

\usepackage{microtype}

\usepackage{inconsolata}

\usepackage{graphicx}

\usepackage{tikz}

\usepackage{colortbl}
\usepackage{amsmath}
\usepackage{array}
\usepackage{tikz}
\usepackage{booktabs}
\usepackage{graphicx}
\usepackage{caption}
\usepackage{subcaption}
\usepackage{subcaption} 
\newcommand{\inc}[1]{\tikz[baseline=(X.base)] 
  \node (X) [inner sep=0pt, anchor=base] {\scriptsize (#1) \textcolor{black}{$\uparrow$}};}

\usepackage[utf8]{inputenc} % allow utf-8 input
\usepackage[T1]{fontenc}    % use 8-bit T1 fonts
\usepackage{hyperref}       % hyperlinks
\usepackage{url}            % simple URL typesetting
\usepackage{booktabs}       % professional-quality tables
\usepackage{amsfonts}       % blackboard math symbols
\usepackage{nicefrac}       % compact symbols for 1/2, etc.
\usepackage{microtype}      
\usepackage{graphicx}

\usepackage{cuted} % 这个包用于创建一个跨两列的环境
\usepackage{capt-of} % 这个包用于使用\captionof命令

\usepackage{xcolor}

\title{Difficult Task Yes but Simple Task No: Unveiling the Laziness in Multimodal LLMs}

\author[1]{Sihang Zhao}
\author[2]{Youliang Yuan}
\author[1]{Xiaoying Tang}
\author[2]{Pinjia He\setcounter{footnote}{1}\thanks{Pinjia He is the corresponding author.}}

\affil[1]{School of Science and Engineering, The Chinese University of Hong Kong, Shenzhen}
\affil[2]{School of Data Science, The Chinese University of Hong Kong, Shenzhen}
\affil[ ]{\texttt{\{sihangzhao, youliangyuan\}@link.cuhk.edu.cn} \authorcr \texttt{\{tangxiaoying, hepinjia\}@cuhk.edu.cn}}

\begin{document}
\maketitle
 
\begin{figure*}[htbp]
  \centering
  \includegraphics[width=1\textwidth]{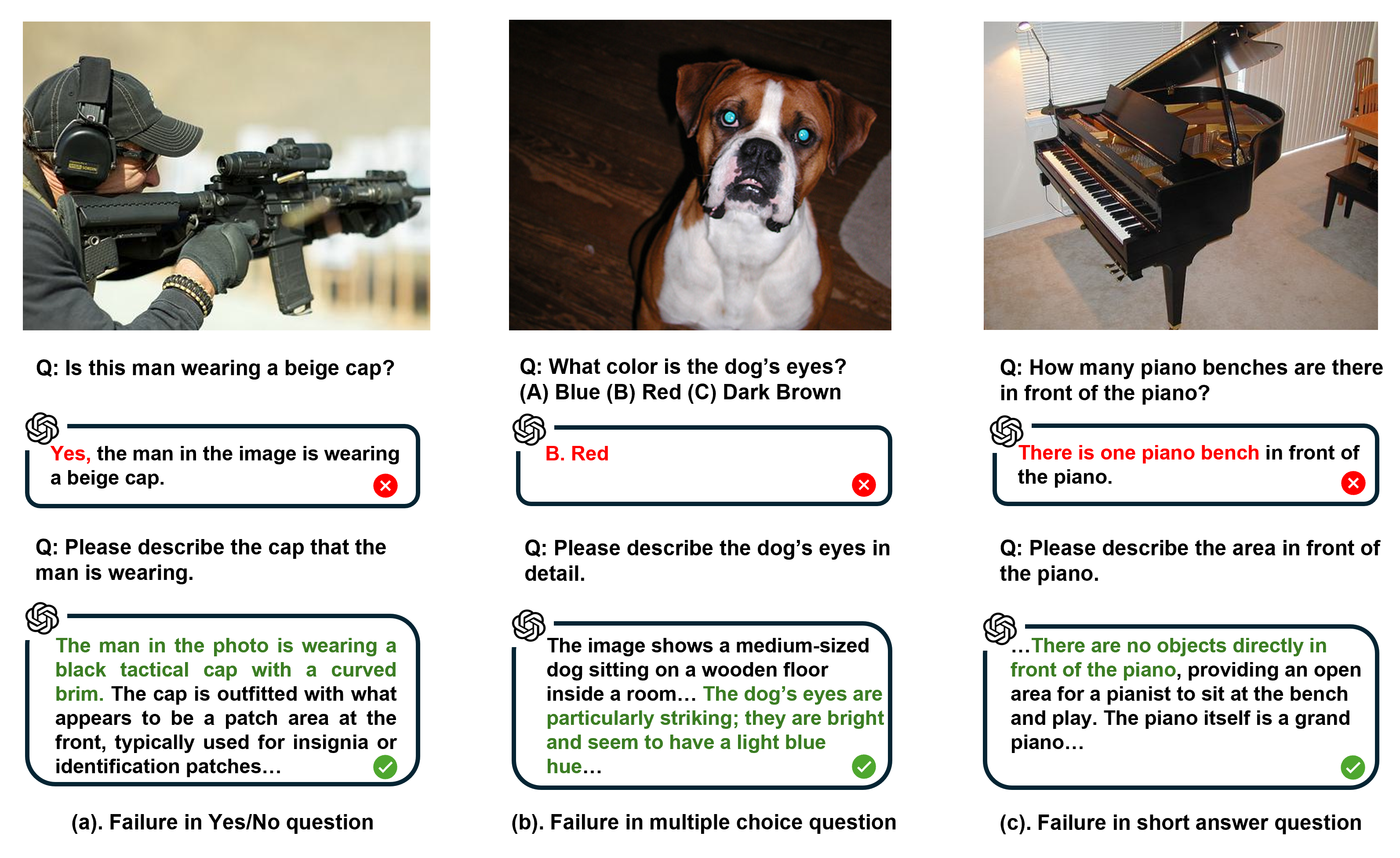}
  \caption{MLLMs sometimes fail to correctly answer straightforward Yes/No or multiple-choice questions based on images. However, they often manage to avoid these errors when describing the images. We refer to this phenomenon as ``model laziness.''}
  \label{fig: examples for MLLMs failures}
\end{figure*}

\begin{abstract}
Multimodal Large Language Models (MLLMs) demonstrate a strong understanding of the real world and can even handle complex tasks. However, they still fail on some straightforward visual question-answering (VQA) problems. 
This paper dives deeper into this issue, revealing that models tend to err when answering easy questions (e.g., Yes/No questions) about an image, even though they can correctly describe it.
We refer to this model behavior discrepancy between difficult and simple questions as model \textit{laziness}.
To systematically investigate model laziness, we manually construct  \textit{LazyBench}, a benchmark that includes Yes/No, multiple choice, short answer questions, and image description tasks that are related to the same subjects in the images.
Based on \textit{LazyBench}, we observe that laziness widely exists in current advanced MLLMs (e.g., GPT-4o, Gemini-1.5-pro, Claude 3, LLaVA-1.5, LLaVA-1.6, and QWen-VL). 
We also analyzed the failure cases of LLaVA-1.5-13B on the VQA-v2 benchmark and discovered that about half of these failures are due to the model’s laziness. This further highlights the importance of ensuring that the model fully utilizes its capability.
To this end, we conduct a preliminary exploration of how to mitigate laziness and find that chain of thought can effectively avoid this issue. The data can be accessed at \href{https://github.com/Akutagawa1998/LazyBench}{https://github.com/Akutagawa1998/LazyBench}.
\end{abstract}

\section{Introduction}
Multimodal Large Language Models (MLLMs) \cite{liu2023llava} integrate multimodal content such as images into large language models (LLMs) \cite{touvron2023llama}. Represented by OpenAI's GPT-4 \cite{openai2023gptsys}, MLLMs have demonstrated impressive capabilities across various complex multimodal tasks \cite{openai2023gpt,yang2023dawn}. However, existing research indicates that even state-of-the-art MLLMs still suffer from some straightforward visual questions (e.g., ``Is the door of the truck cab open?'' for an image of a truck.) \cite{tong2024eyes}.
A natural question arises:
\begin{quote}
    \it Why do MLLMs struggle with these easy questions? 
    % Can MLLMs understand these simple images in failure cases? 
\end{quote}

In this work, we dive deeper to explore this question and find that  MLLMs often struggle with simple questions (like Yes/No questions) about an image, even though they can accurately describe the image itself.
For example, as present in Figure \ref{fig: examples for MLLMs failures}, when we asked GPT-4V, ``\texttt{Is this man wearing a \textit{beige} cap?}'' GPT-4V answered ``\texttt{Yes}'', which is incorrect. In contrast, when we asked it a similar but more difficult question, ``\texttt{Please describe the cap that the man is wearing}'', GPT-4V correctly described its color. 
In this paper, we describe this phenomenon where MLLMs perform well on the description tasks but make mistakes on simpler tasks as model \textbf{laziness}\footnote{The explanation of why we define description tasks as the harder tasks can be found in Appendix \ref{why hard}}.
%We think GPT-4V can understand the picture but it fails on simple Yes/No questions because of its being \textit{lazy}. 
% Instead, it is "being lazy" when faced with "simpler" Yes/No questions. 
% We further observed that this phenomenon is widespread among other state-of-the-art MLLMs: they can often perform poorly on some simple VQA tasks that only require "Yes/No" answers or multiple-choice responses. However, when we ask the models to describe the image of the corresponding subject in the original questions, the models tend to give relatively more decent responses.

% We define the above behaviour as MLLMs' \textit{laziness} in this study, namely, the model correctly describes the detail of the image but fails to correctly respond with Yes/No, or multiple choice questions about the same statement.

To systematically study model laziness, we manually construct a benchmark called \textbf{LazyBench}. We found image pairs encoded as ``similar images'' by the pretrained Contrastive Language-Image Pre-Training (CLIP) \cite{radford2021learning} model and designed simple Yes/No questions on their visual differences. We collected images from where GPT-4V (website version) fails in the above-mentioned questions. Then for each image, we handcraft three different types of questions about the same subject of the Yes/No question: multiple choice, short answer question, and a description task.
We use LazyBench to evaluate advanced closed-source models like GPT-4o, GPT-4V \cite{openai2023gpt}, Gemini-1.5-pro \cite{reid2024gemini}, and Claude 3 \cite{claude3}, and open-source models like LLaVA-1.5 \cite{liu2023llava}, LLaVA-1.6 \cite{liu2024visual} and QWen-VL \cite{Qwen-VL}. 
The results show that these state-of-the-art MLLMs significantly suffer from laziness: they show a low accuracy on Yes/No questions (e.g., GPT-4V: 28.72\%, Claude 3: 34.66\%), and multiple choice questions (e.g., GPT-4V: 54.45\%, Claude 3: 55.45\%), while performing significantly better on the corresponding description tasks (e.g., GPT-4V: 71.28\%, Claude 3: 57.43\%).

We further explore to what extent MLLM laziness is prevalent in the widely used visual question-answering (VQA) benchmarks. 
To this end, we propose a simple LLM-based framework that automatically evaluate  es the extent of laziness in their failure cases. 
 
We find that 41.15\% failure cases of LLaVA-1.5-13B on VQA-v2 are caused by model laziness. 
We believe this provides valuable insights into the way to improve the capability of MLLMs: in addition to allowing MLLMs to learn more knowledge, it is equally important to ensure that MLLMs are fully utilizing the knowledge learned.

To mitigate the influence caused by model laziness in simple tasks, we implemented a chain of thought (CoT) \cite{wei2022chain} based method to make the task ``harder''. We require MLLMs to handle the description task first before answering a Yes/No or a multiple-choice question. 
The results show that our method fixed around 40\% cases of laziness and effectively improved MLLMs' performance in those tasks.

%We use both GPT-4-turbo and human evaluator to evaluate whether the model descriptions match the ground truth statements of the images. 

In summary, our contributions are listed below:
\begin{itemize}
    \item We conduct an in-depth study on the phenomenon of MLLMs making errors on easy questions, discovering that current advanced MLLMs exhibit significant laziness.

    \item We manually construct a dataset called LazyBench to investigate the laziness phenomenon in MLLMs.

    \item We provide a CoT-based method that can effectively prevent models from being lazy.

\end{itemize}

\begin{figure*}[h]
  \centering
  \includegraphics[width=0.95\textwidth]{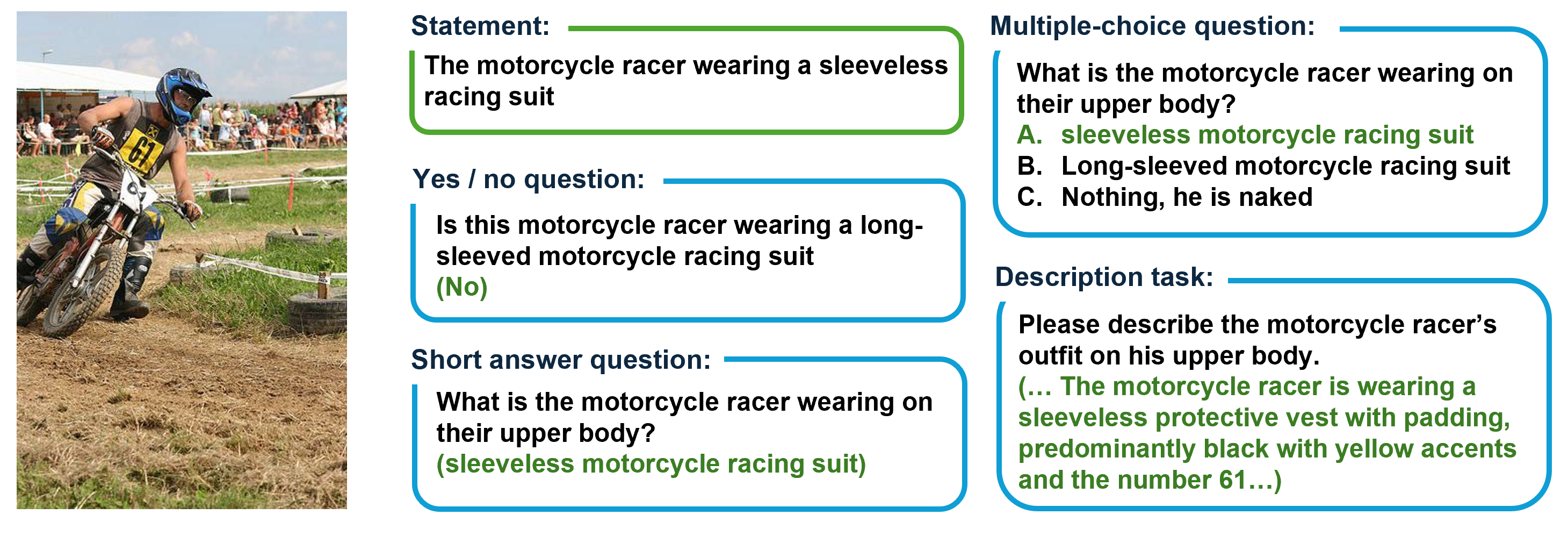}
  \caption{The green box represents a correct, brief statement about the ``question subject'' in the image. The blue box contains four different types of questions about this subject (Yes/No, multiple-choice, short-answer questions, and descriptive requests). They are used to evaluate the model's laziness, and the construction of these questions is described in Section \ref{construction}. }
  \label{fig: example of Lazybench}
\end{figure*}

% \begin{figure*}
%   \centering
%   \includegraphics[width=1\textwidth]{images/construct_lazybench.png}
%   \caption{The Process of constructing \textbf{LazyBench}: we utilize CLIP \cite{radford2021learning} to identify images that the model considers "similar" and analyze the differences between them to pinpoint instances where MLLMs provide incorrect answers. Based on these errors, we construct a series of related questions. }
%   \label{fig: construct lazy}
% \end{figure*}

\section{Related Work}
\label{related work}
\subsection{Visual Question and Answering}
\label{vqa}
With the success of LLMs, increasing attention has been given to integrating visual embeddings into language models. Initially, researchers applied transformers to connect visual encoders with LLMs, pretraining them on image-text matching datasets \cite{lin2014microsoft, krishna2017visual, changpinyo2021conceptual} and fine-tuning them on specific datasets (e.g., VQA \cite{antol2015vqa}, VQA-v2 \cite{goyal2017making}). Then, to improve MLLMs' performances and generalization abilities, researchers began using VQA format data for instruction tuning \cite{liu2023llava}. Despite MLLMs showing considerable capabilities in some complex VQA tasks \cite{fu2022there, hu2022promptcap, hu2023tifa, fu2023generate, fu2023interpretable}, these studies seem to focus primarily on the textual reasoning abilities of MLLMs \cite{wei2022chain}, rather than on whether MLLMs are truly extracting information from the images. 
Our work bridges this gap by studying the model laziness.

\subsection{Benchmarks for Visual Perceptions}
\label{benchmarks}
Increasing attention is being given to the evaluation of MLLMs' visual perception. \citealp{tong2024eyes} suggest that due to encoding flaws in the CLIP pre-trained model, CLIP-based MLLMs might make mistakes on some simple questions. POPE \cite{li2023evaluating} and NOPE \cite{lovenia2023negative}  designed questions about the presence or absence of objects in images to measure MLLM hallucination; however, these consist solely of Yes/No questions. Hallusibench \cite{liu2023hallusionbench} provides a benchmark for evaluating MLLMs' hallucinations across different tasks. MathVerse \cite{zhang2024mathverse} is a benchmark for visual problems in mathematical domains such as tables and charts. It reveals that MLLMs may not be thoroughly reading these charts, but they lack analysis of simpler and more straightforward VQA tasks. LazyBench is the first benchmark to focus on the consistency of MLLMs' answers to the same question about the same subject in the same image when asked in different forms.

\section{MLLMs Are Being Lazy}
\label{mllms are lazy}

To thoroughly understand and analyze the lazy phenomenon, where MLLMs perform well on descriptive tasks but fail on simple tasks, we construct the \textbf{LazyBench} benchmark. Therefore, in this section, we first introduce the methods and steps for constructing LazyBench. Subsequently, we measure the extent of the lazy phenomenon of current state-of-the-art MLLMs on LazyBench. Finally, we use a CoT-based method to mitigate the MLLMs laziness.

\subsection{Samples of LazyBench}
% \begin{figure*}[h]
%   \centering
%   \includegraphics[width=0.95\textwidth]{images/sample_lazybench.png}
%   \caption{The green box represents a correct, brief statement about the "question subject" in the image. The blue box contains four different types of questions about this subject (Yes/No, multiple-choice, short-answer questions, and descriptive requests). They are used to evaluate the model's laziness, and the construction of these questions is described in Section \ref{construction} }
%   \label{fig: example of Lazybench}
% \end{figure*}

\begin{figure*}
  \centering
  \includegraphics[width=1\textwidth]{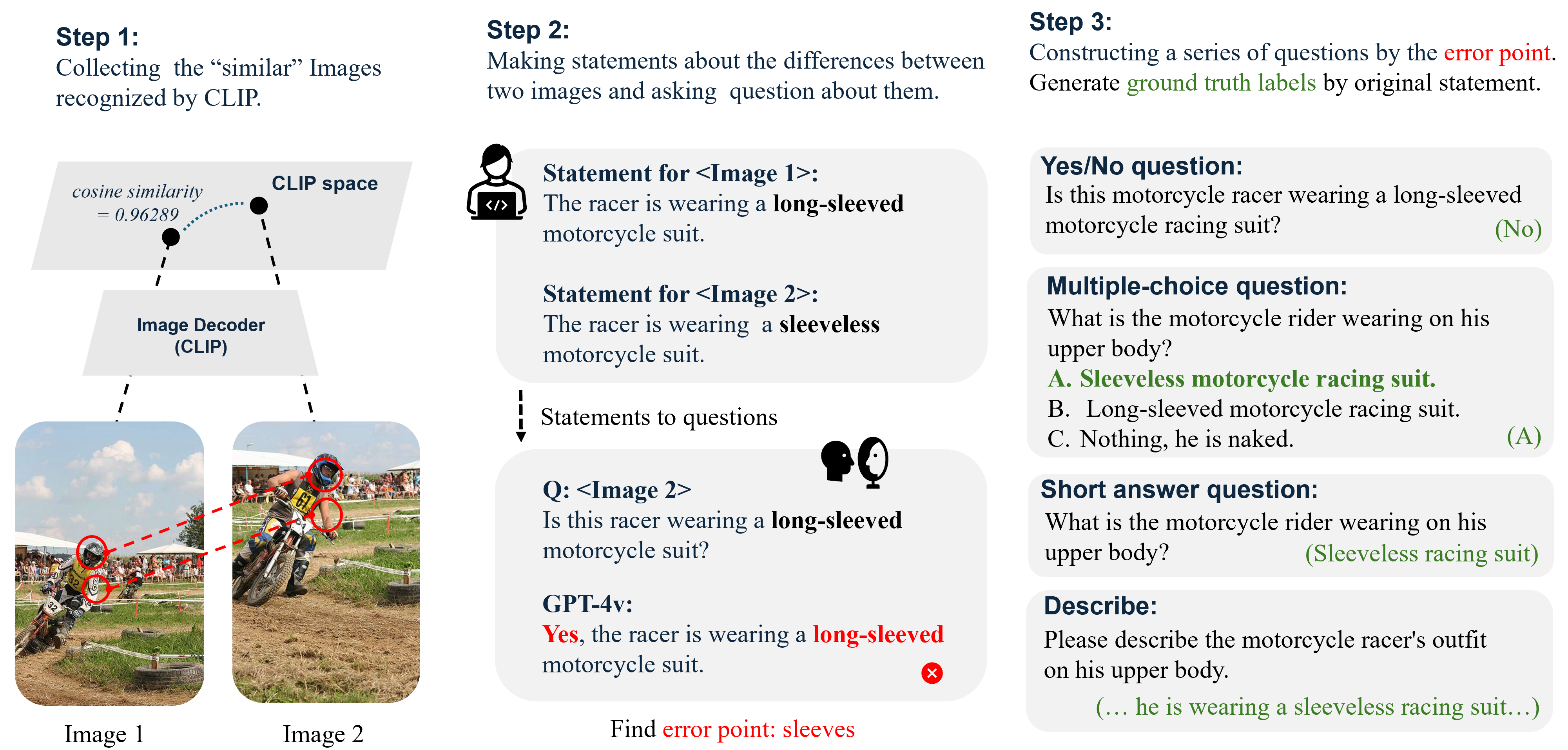}
  \caption{The Process of constructing \textbf{LazyBench}: we utilize CLIP \cite{radford2021learning} to identify images that the model considers ``similar'' and analyze the differences between them to pinpoint instances where MLLMs provide incorrect answers. Based on these errors, we construct a series of related questions. }
  \label{fig: construct lazy}
\end{figure*}

Each item in \textbf{LazyBench} consists of an image, a ground truth statement, and 4 different questions (i.e., Yes/No, multiple-choice, short answer, description) together with their ground truth answers. For instance, in Figure \ref{fig: example of Lazybench}, for the image, there are:

\begin{itemize}
    \item One Yes/No question: ``Is this motorcycle racer wearing a long-sleeved motorcycle racing suit?'' and its ground truth answer is ``No''.

    \item One multiple-choice question has 3 options: ``sleeveless motorcycle racing suit'', ``Long-sleeved motorcycle racing suit'', ``Nothing, he is naked'' and the first one as its ground truth.

    \item One short answer question: ``What is the motorcycle racer wearing on their upper body?''

    \item One description question: ``Please describe the motorcycle racer's outfit on his upper body.'' which is an open-ended question and should be related to the statement.
\end{itemize}

In total, we have 101 images and 404 questions together with their ground truth labels and distract options.

\subsection{Constructing LazyBench}
\label{construction}

% \textbf{Data Source}: We reuse the images from ImageNet (ILSVRC2012)\cite{ILSVRC15} and MMVP\cite{tong2024eyes}. \yyl{add the reason why we do like this}

Inspired by \citealp{tong2024eyes}, we designed a series of questions targeting the visually obvious differences between image pairs that were encoded as similar by CLIP \cite{radford2021learning}. Intuitively, if two images are encoded as similar vectors by CLIP but have clear visual differences, it indicates that at least one of the images had certain features incorrectly encoded or neglected. This step helps us quickly construct a set of visual questions that MLLMs are likely to get wrong. We collected images from ImageNet \cite{russakovsky2015imagenet} and MMVP \cite{tong2024eyes}. The specific steps are listed below:

\textbf{Image Selection} We encoded each image using CLIP and compared their cosine similarities. Followed by \citealp{tong2024eyes} and our observation, here we focused on ``similar image pairs'' with a cosine similarity greater than 0.96 but smaller than 0.99. This similarity ensures that the images in the pair are considered ``very similar'' by CLIP yet also easy to find obvious visual differences between the image pairs. We then identified images that appeared significantly different in human view.

\textbf{Question Construction} Based on images from the previous step, we formulated Yes/No questions targeting their differences. We collected the images and questions that might be answered incorrectly\footnote{We use the web version GPT-4V as the filter.}  and designed ground truth statements, multiple-choice questions, short answer questions, and descriptive request questions around the error points. The process is shown in Figure \ref{fig: construct lazy}. The ground truth of Yes/No questions will always be ``no'' and the correct option for multiple-choice questions is shuffled randomly in A, B, and C. % We designed a pilot study to ensure that models are not answering "yes" without hesitation. The details of this pilot study are in the Section 3.4.

\begin{table*}[htbp]
\centering
\setlength{\tabcolsep}{3pt}
\caption{Evaluation result for MLLMs on LazyBench. Underline indicates in which task this model performs best and bold denotes the model that gives the best performance in this task.}
\begin{tabular}{lccccccc}
% \multicolumn{7}{c}{\textbf{Evaluation Result for MLLMs on LAZY Benchmark}}                       \\ 
\toprule
\multirow{2}{*}{Model}        & \multicolumn{2}{c}{Yes/No}  & \multicolumn{2}{c}{Multiple Choice}  & \multicolumn{2}{c}{Short Answer} & Description \\ 
% \cmidrule(lr){2-3} \cmidrule(lr){4-5}
      & Accuracy & Lazy Rate          & Accuracy & Lazy Rate                  &Accuracy & Lazy Rate             & Accuracy            \\ \hline
GPT-4o      & \textbf{60.40}  & 75.00           & \textbf{78.22}  & 37.50                    & \textbf{69.37} &58.06        & \textbf{\underline{84.16}}                \\
GPT-4V      & 28.72  & 70.83           & 54.45  & 37.50                    & 55.33&48.89       & \underline{69.77}                \\
Gemini-1.5-pro & 50.50  & 70.00           & 62.38  & 46.00                    & 58.42&50.00        & \underline{76.24}                \\
Claude 3     & 34.65  & 62.12           & 54.45  & 42.42                    & 48.51&38.09       & \underline{59.34}                \\
$\text{LLaVA-1.5}^{13B}$ & 34.65  & 53.03           & \underline{52.48} & 25.75            & 45.54&54.46      & 48.51\    \\
$\text{LLaVA-1.6-Mistral}^{7B}$     &35.64      &66.15          &57.43            &36.92              &47.52              &47.17              &\underline{67.33}\\
$\text{LLaVA-1.6-Vicuna}^{7B}$     &32.67      &55.88          &49.50             &35.29              &46.53              &43.40              &\underline{55.54}\\
$\text{LLaVA-1.6-Vicuna}^{13B}$     &35.64      &73.85          &57.43             &43.08              &46.53              &62.26              &\underline{70.30}\\
$\text{LLaVA-1.6-Vicuna}^{34B}$     &56.44      &59.90          &\underline{66.34}             &25.00             &55.43              &34.15              &65.35\\
Qwen-VL-Plus    &49.50      &58.82          &59.41             &37.25              &53.47              &44.68              &\underline{66.34}\\
Qwen-VL-Max     &47.52      &58.49          &64.36             &28.30              &60.40              &49.90              &\underline{69.31}\\
% Average     &41.70      &66.20          &60.40             &37.83              &55.43              &37.50              &\underline{67.60}\\
% LLaVA-1.5-7B &  0.00 &  - & 33.66 &  - & \underline{38.61} & \underline{38.61}     \\

\bottomrule
\end{tabular}
\label{table: table 1}
\end{table*}

When designing the description request, we directly asked the model to describe the subject of our focus (e.g., in Figure \ref{fig: example of Lazybench}, we requested the model to describe the motorcycle racer's outfit on his upper body). This means that the subject of the Yes/No questions and multiple-choice questions was equivalently addressed in the description request. So the description request does not include any additional information or prompt any CoT guidance.

\subsection{Experimental Result}
\label{experimental result}

\paragraph{Setup} For evaluating model laziness, we assessed the LazyBench questions on SOTA close-source MLLMs such as GPT-4o, GPT-4-Vision-preview \cite{openai2023gpt}, Gemini-1.5-pro \cite{reid2024gemini}, Claude-3-Opus-20240229 \cite{claude3} and the open-source model LLaVA-1.5\footnote{We found LLaVA-1.5-7B tends to answer ``yes'' for all Yes/No questions, therefore, we used the 13B version only. The detailed evaluation can be found in Appendix \ref{addition evaluation}.} \cite{liu2023llava}, LLaVA-1.6-Mistral-7B, LLaVA-1.6-Vicuna (7B, 13B, 34B) \cite{liu2024visual}, QWen-VL-Plus and QWen-VL-Max \cite{Qwen-VL}. We set the temperature to 0 to make our results reproducible. 

\paragraph{Evaluation} We classified instances where models made errors on Yes/No, multiple-choice or short answer questions but provided accurate descriptions of the related image as instances of ``being lazy''. We defined ``lazy rate'' as the number of lazy cases divided by the number of total failure cases on the simpler questions. We used a binary classification to score the descriptions provided by MLLMs as either correct (1) or incorrect (0). the detailed evaluation criterion can be found in Appendix \ref{Evaluation Criterion}.

\begin{figure*}[h]
  \centering
  \includegraphics[width=1\textwidth]{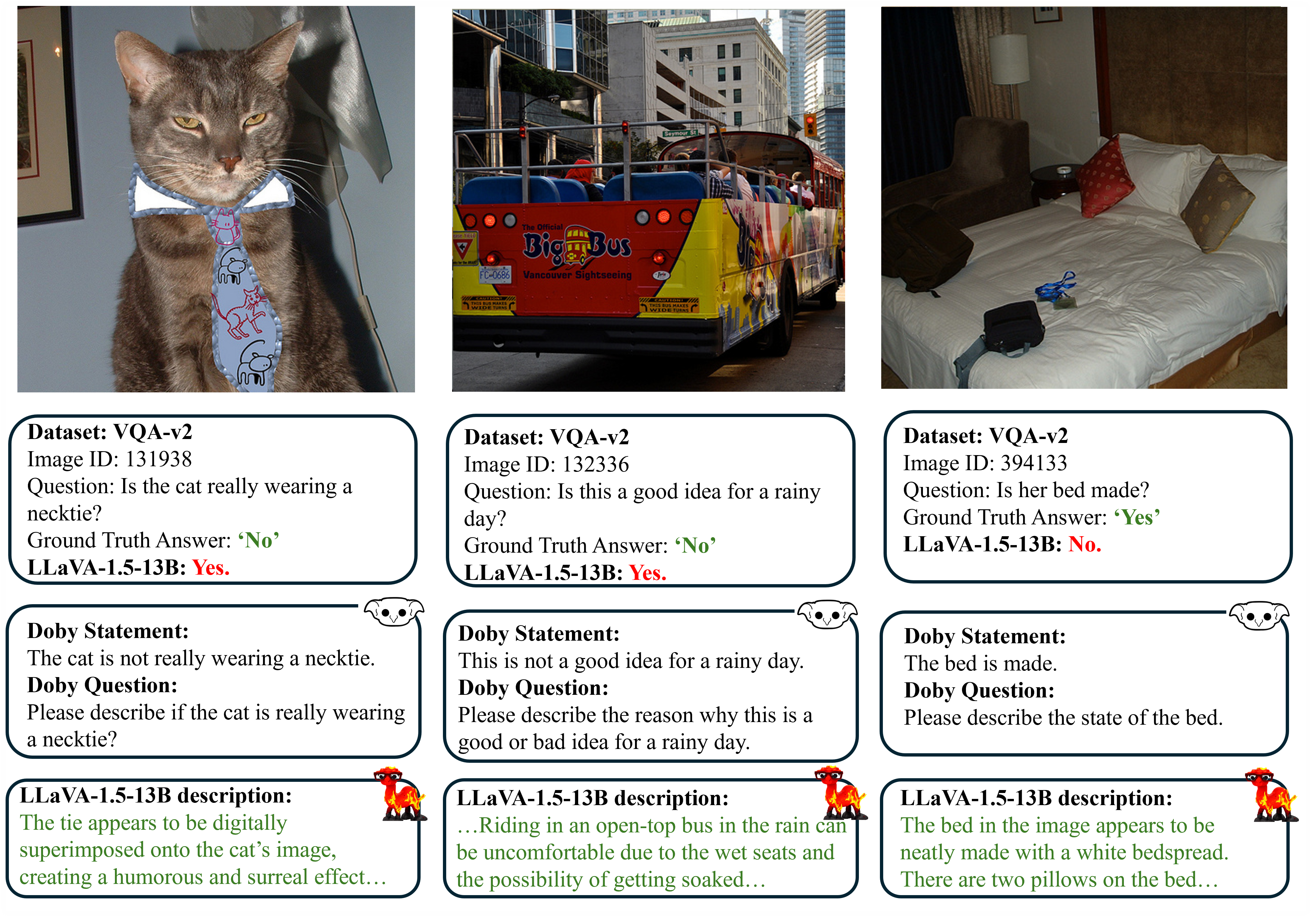}
  \caption{Examples of LLaVA-1.5-13B being lazy in VQA-v2. The first line of boxes below each image contains the original labels and questions in VQA-v2, as well as the initial responses from LLaVA-1.5-13B. The second line of boxes contains the statement and description request automatically generated by Doby. The last line contains the responses of LLaVA-1.5-13B to Doby's questions. Subsequently, by comparing these responses to the statement, it is determined whether the model is being lazy in these cases.}
  \label{fig: example_of_doby}
\end{figure*}

\begin{table*}[htbp]
\centering
\caption{The performance improvement of MLLMs on LazyBench after the CoT-based prompt. Bold donates in which task this model performed best, Smaller numbers in the `Accuracy' columns denote the improvement compared to the accuracy that MLLMs directly answer the questions. (Fix Rate: the proportion of laziness cases that have been fixed.)}
\label{table:performance_comparison}
\begin{tabular}{lcccccc}
% \multicolumn{7}{c}{\textbf{Evaluation Result for MLLMs on LAZY Benchmark}}                       \\ 
\toprule
% models & Yes/No &  Yes/No (CoT) & multiple choice & multiple choice (CoT)& Description  \\ \midrule
\multirow{2}{*}{Model}            & \multicolumn{2}{c}{Yes/No}  & \multicolumn{2}{c}{Multiple Choice} & Description \\ 
    & Fix Rate & Accuracy           & Fix Rate & Accuracy  & Accuracy &             \\ \hline
GPT-4o              & 37.50 & \cellcolor{green!100}71.29\inc{24.76}  & 43.48 & \cellcolor{green!30}\textbf{84.16}\inc{5.94} & \textbf{84.16} \\
GPT-4V             & 41.67 & \cellcolor{green!100}52.48\inc{23.76}  & 47.92 & \cellcolor{green!50}66.34\inc{11.69} & \textbf{71.28} \\
Gemini-1.5-pro      & 44.00 & \cellcolor{green!60}64.36\inc{13.86}   & 26.82 & \cellcolor{green!30}67.33\inc{4.95} & \textbf{76.24} \\
Claude-3            & 40.91 & \cellcolor{green!80}52.48\inc{19.81}   & 42.11 & \cellcolor{green!30}\textbf{58.42}\inc{3.96} & 57.43 \\
$\text{LLaVA-1.5}^{13B}$      & 36.36 & \cellcolor{green!60}50.50\inc{15.58}   & 54.55 &\cellcolor{green!10}\textbf{53.47}\inc{0.99} & 48.51 \\ 
\bottomrule
\end{tabular}
\end{table*}
\textbf{MLLMs are being lazy on over 50\% failures in Yes/No questions}. 
As the result shown in Table \ref{table: table 1}, most of the MLLMs perform their best on description tasks and have the worst responses on Yes/No questions. Specifically, on GPT-4V, the accuracy for Yes/No questions is less than 30\%, while the accuracy improves for multiple-choice and short-answer questions. In description tasks, GPT-4V achieves an accuracy of 69.77\%, which is 41.05\% higher than Yes/No questions and 15.32\% higher than multiple-choice questions. This indicates that GPT-4V indeed exhibits laziness when facing ``simple tasks.'' Similar results can also be observed with Claude 3, where the accuracy for Yes/No questions was only 34.66\%, while the accuracy for descriptions reached 59.34\% and other MLLMs in the table.

\textbf{Strong closed-source models tend to exhibit high lazy rates:}
We found that all closed-source models exhibit an over 60\% lazy rate on Yes/No questions. For multiple-choice questions, the ``lazy rate'' for all closed-source models exceeds 35\%. The top two best-performing models in our evaluation are GPT-4o and Gemini-1.5-pro. They achieved 60.4\% and 50.5\% accuracy on Yes/No questions. GPT-4o attained a multiple-choice accuracy of 78.22\% and a description accuracy of 84.16\%, while Gemini-1.5-pro reached 62.35\% and 76.24\%. However, they also exhibit the most severe lazy rate on these tasks (GPT-4o: 75\% in Yes/No questions, Gemini-1.5-pro: 46\% in multiple choice questions.) This indicates that despite improvements in model capabilities, the phenomenon of MLLMs laziness persists, even stronger.

\section{Discussion}
\subsection{Laziness in Existing Benchmarks}
To explore the impact of laziness on the evaluation of MLLMs in existing benchmarks for visual perceptions, we conducted case studies on several popular benchmarks (e.g., VQA-v2 \cite{goyal2017making} and Hallusionbench \cite{liu2023hallusionbench}). We evaluate LLaVA-1.5-13B on 1000 Yes/No questions in the VQA-v2 validation set. To automate this process, we design and propose \textbf{Do}n't \textbf{b}e laz\textbf{y} (\textbf{Doby}). Doby is a framework based on GPT-4o which can generate the ground truth statements and description requests like LazyBench from the Yes/No or multiple choice questions-answering pairs in the existing datasets, thereby expanding the original datasets. After the MLLMs respond to the descriptive tasks, Doby compares the generated statements with the model's descriptions to determine if the tested MLLMs' descriptions accurately convey the relevant information. This process allows for automatic monitoring and statistical analysis of the MLLMs' laziness phenomenon among the datasets.

Using Doby, we found that LLaVA-1.5-13B is lazy in 79 of 192 failure cases. Some examples are given in Figure \ref{fig: example_of_doby}. This indicates that the model's inability to sufficiently utilize internal knowledge under simple tasks is also an important reason for the model's insufficient accuracy. Namely, how to prevent the model from being lazy is an important part of improving the model's capabilities.

\subsection{Mitigating Laziness}

As the above experimental results show that MLLMs are lazy in the simpler tasks. We also want to know:
\begin{quote}
    \it Can we mitigate this phenomenon by making the questions harder? 
    % Can MLLMs understand these simple images in failure cases? 
\end{quote}
To answer this question, we used a CoT-based method that let MLLMs answer the Yes/No and multiple choice questions after letting them finish the description task. For example, we ask the MLLMs about the image shown in Figure \ref{fig: example of Lazybench}: Please describe the motorcycle racer's outfit on his upper body, and then answer the question: \textit{Is this motorcycle racer wearing a long-sleeved motorcycle racing suit?}

As Table \ref{table:performance_comparison} shows, after employing this CoT method, MLLMs exhibit significant improvements in both Yes/No and multiple-choice questions. The enhancement is more pronounced for Yes/No questions. Among the models, GPT-4o, which had the highest accuracy in description tasks, showed the greatest improvement in Yes/No questions. Specifically, GPT-4o's accuracy in Yes/No questions increases by 24.76\%. There are 37.5\% GPT-4o laziness cases among the original Yes/No questions that have been repaired, while Gemini-1.5-pro and LLaVA see the least improvements of 13.86\% and 15.58\%. Additionally, GPT-4o's accuracy in multiple-choice questions improves by 5.94\%, matching its performance in description tasks, while the accuracy for Claude 3 and LLaVA-1.5-13B even slightly exceeds their performance in description tasks.

We further hypothesize that fine-tuning MLLMs to provide explanations before giving answers, rather than answering first and then explaining \cite{chu2024better}, could also reduce MLLMs' laziness. Similarly, the method proposed by \citealp{yuan2024refuse} allows models to correct themselves while generating unsafe outputs, which might also be effective in this context: when MLLMs realize that their first one or few tokens (e.g., ``Yes'', ``A'', etc.) of their initial answer may have been incorrect while explaining, they can adjust and improve their response. The automatic prompt may also be useful \cite{pryzant2023automatic}.

\subsection{Doby Helps Find Noise Sample}
Furthermore, by checking the response to description request of Doby, we find that in addition to instances of laziness (Figure \ref{fig:fig 5} in Appendix \ref{sec:hallusionbench}), these datasets contain numerous issues like the textual information of the question is vague (Figure \ref{fig:fig 5}(d)), or the questions cannot be answered solely based on the images (Figure \ref{fig:fig 5}(c)). Ignoring these issues may lead to incorrect assessments of the model's capabilities. These issues are not apparent when solely examining the results of MLLMs on Yes/No questions and multiple choice questions, which also suggests that future researchers should take a deeper look into the description response.

\subsection{Further Discussion}
\label{addition evaluation}

As previous studies \cite{hu2023ciem,liu2023aligning} have found imbalanced training data often causes many MLLMs to directly give affirmative answers like ``yes'' to any question. To further verify that MLLMs' laziness is different from option bias, we construct the conversed statement by another image in the ``similar image pairs'', (e.g., ``Statement for Image\_1'' in Step 2 of Figure \ref{fig: construct lazy}). The detailed information can be found in Appendix \ref{sec:example of conv}.

\begin{table}[htbp]
\centering
\caption{Accuracy of Irrelevant Questions (All answers are ``No'') and Conversed Question (All answers are ``Yes''.)}
\begin{tabular}{lcc}
\toprule
Model  & Irrelevant & Conversed \\ \midrule
GPT-4o & 96.04 & 81.19 \\ 
GPT-4V & 92.07 & 64.35 \\ 
Gemini-1.5-pro & 90.10 & 64.36 \\ 
Claude 3 & 88.11 & 57.42 \\ 
$\text{LLaVA-1.5}^{13B}$ & 67.32& 80.20 \\ 
$\text{LLaVA-1.6-Mistral}^{7B}$     &73.27      &77.23 \\
$\text{LLaVA-1.6-Vicuna}^{7B}$     &76.24       &77.23\\
$\text{LLaVA-1.6-Vicuna}^{13B}$     &80.20      &76.24\\
$\text{LLaVA-1.6-Vicuna}^{34B}$     &94.06      &52.48\\
Qwen-VL-Plus    &87.13      &79.21\\
Qwen-VL-Max     &89.11      &68.32\\
\cellcolor{gray!30}$\text{LLaVA-1.5}^{7B}$ & \cellcolor{gray!30}0.00 
&\cellcolor{gray!30}100.00\\ 
\bottomrule
\end{tabular}
\label{table:table_irr}
\end{table}

\begin{table*}[htbp]
\centering
\caption{Evaluation Result for MLLMs on LazyBench (Rev Rate: the proportion that MLLMs give incorrect responses to the description task but successfully give the correct answer to Yes/No.)}
\begin{tabular}{lccccccc}
% \multicolumn{7}{c}{\textbf{Evaluation Result for MLLMs on LAZY Benchmark}}                       \\ 
\toprule
  \multirow{2}{*}{Model}          & \multicolumn{3}{c}{Yes/No}    & Description \\ 
      & Accuracy & Lazy Rate & Rev Rate                        & Accuracy            \\ \hline
GPT-4o      & 60.40  & 75.00  & 37.50                      & 84.16                \\
GPT-4V      & 28.72  & 70.83  & 25.00                           & 69.77              \\
Gemini-1.5-pro & 50.50  & 70.00 &37.50                     & 76.24               \\
Claude 3     & 34.65  & 62.12  & 30.55                        & 59.34                \\
$\text{LLaVA-1.5}^{13B}$ & 34.65  & 53.03 & 32.61                    & 48.51    \\ 
$\text{LLaVA-1.6-Mistral}^{7B}$     &35.64      &66.15          &33.33     &67.33\\
$\text{LLaVA-1.6-Vicuna}^{7B}$     &32.67      &55.88          &33.33      &55.54\\
$\text{LLaVA-1.6-Vicuna}^{13B}$     &35.64      &73.85          &43.33     &70.30\\
$\text{LLaVA-1.6-Vicuna}^{34B}$     &56.44      &59.90          &48.57     &65.35\\
Qwen-VL-Plus    &49.50      &58.82          &28.24             &66.34\\
Qwen-VL-Max     &47.52      &58.49          &29.03             &69.31\\
%LLaVA-1.5-7B &  0.00 &  - & -  & \textbf{38.61}     \\

\bottomrule
\end{tabular}
\label{table: rev}
\end{table*}

In the open-source model LLaVA-1.5-7B, the laziness seems not as apparent as in the closed-source models. We found LLaVA-1.5-7B exhibits severe bias in Yes/No questions and answers ``yes'' for all questions, as shown in Table \ref{table:table_irr}. 
This explains its performance (i.e., a random guessing accuracy 33.66\%) in multiple-choice questions. So we do not consider LLaVA-1.5-7B when analysing the MLLMs' laziness. As the model size increases, the tendency of LLaVA-1.5-13B to ``thoughtlessly'' answer ``yes'' to Yes/No questions is significantly alleviated. The closed-source MLLMs also have decent performances in these questions. The result shows this option bias is different from MLLMs' laziness.

In previous experiments, we mainly focused on the lazy rate, which refers to cases where the model answers Yes/No questions incorrectly but correctly describes the scenarios. To further validate our findings, we answer the question below: \textit{can we also find a significant number of cases where the model makes mistakes in descriptions but answers Yes/No questions correctly?}

It is intuitive and normal to make mistakes on more difficult tasks and perform well on simpler ones, the small label space of Yes/No tasks means that even random guessing has a 50\% chance of being correct. In Table \ref{table: rev}, we provide the results regarding Rev Rate (i.e., the proportion that MLLMs give incorrect responses to the description but successfully give the correct answer to Yes/No). The results show that the rev rate is significantly lower than the lazy rate. Considering that Yes/No questions are easy to guess while describing questions are hard to answer through guessing, we believe the experimental results answer the above question well: The phenomenon of laziness truly exists.

\subsection{Why the MLLMs are Lazy?}
We have a hypothesis about the reason why MLLMs are lazy: take Yes/No questions and descriptions as examples. For the former, the answer (MLLMs response) needs to be given within a few tokens or even a single token (i.e., ``Yes'', ``No'', or ``A'' etc.), which means the model can only ``look at the image a few times or even just once'' while decoding the answer. In contrast, when generating the description of a specific region in the image, MLLMs may need to look at the image many times throughout the decoding process. The ``quick glance'' for simple tasks versus the ``careful observation'' for complex tasks might be the reason behind laziness. We believe it is important to understand and explain laziness accurately with more experiments. However, since we are in the early stages of studying laziness, in this work we focus more on measuring, understanding its impacts, and finding solutions for laziness. We will leave the in-depth exploration of laziness for the future.

\section{Conclusion}
This paper highlights the \textit{laziness} in MLLMs: a model can handle difficult tasks (e.g., describe the subject) but fails on simple tasks (e.g., a corresponding Yes/No question). We provide a benchmark \textit{LazyBench} that systematically shows this discrepancy in model performance across advanced MLLMs. 
Our findings indicate that in addition to allowing the model to learn more knowledge, it is equally important to ensure that MLLM is fully utilizing the knowledge learned.
% Implementing a CoT intervention successfully avoids this laziness, enhancing model performance notably. These findings suggest that 
% besides expanding the knowledge base, ensuring MLLMs effectively utilize their capabilities is crucial for advancing their practical utility and reliability in real-world applications.

\section*{Limitations}
This paper has the following limitations. First, laziness mainly occurs in powerful closed-source MLLMs where we cannot access their internals for further analysis of the root causes. Second, although our CoT-based method shows preliminary effectiveness, we regard the development and evaluation of laziness mitigation mechanisms as important future work. Third, the size of LazyBench is small. We will keep expanding it in the future. 
\bibliography{acl_latex}

\begin{thebibliography}{32}
\providecommand{\natexlab}[1]{#1}

\bibitem[{{Anthropic}(2024)}]{claude3}
{Anthropic}. 2024.
\newblock Claude3.
\newblock \url{https://www.anthropic.com/claude,}.

\bibitem[{Antol et~al.(2015)Antol, Agrawal, Lu, Mitchell, Batra, Zitnick, and Parikh}]{antol2015vqa}
Stanislaw Antol, Aishwarya Agrawal, Jiasen Lu, Margaret Mitchell, Dhruv Batra, C~Lawrence Zitnick, and Devi Parikh. 2015.
\newblock Vqa: Visual question answering.
\newblock In \emph{Proceedings of the IEEE international conference on computer vision}, pages 2425--2433.

\bibitem[{Bai et~al.(2023)Bai, Bai, Yang, Wang, Tan, Wang, Lin, Zhou, and Zhou}]{Qwen-VL}
Jinze Bai, Shuai Bai, Shusheng Yang, Shijie Wang, Sinan Tan, Peng Wang, Junyang Lin, Chang Zhou, and Jingren Zhou. 2023.
\newblock Qwen-vl: A versatile vision-language model for understanding, localization, text reading, and beyond.
\newblock \emph{arXiv preprint arXiv:2308.12966}.

\bibitem[{Changpinyo et~al.(2021)Changpinyo, Sharma, Ding, and Soricut}]{changpinyo2021conceptual}
Soravit Changpinyo, Piyush Sharma, Nan Ding, and Radu Soricut. 2021.
\newblock Conceptual 12m: Pushing web-scale image-text pre-training to recognize long-tail visual concepts.
\newblock In \emph{Proceedings of the IEEE/CVF conference on computer vision and pattern recognition}, pages 3558--3568.

\bibitem[{Chu et~al.(2024)Chu, Chen, and Nakayama}]{chu2024better}
KuanChao Chu, Yi-Pei Chen, and Hideki Nakayama. 2024.
\newblock A better llm evaluator for text generation: The impact of prompt output sequencing and optimization.
\newblock \emph{arXiv preprint arXiv:2406.09972}.

\bibitem[{Fu et~al.(2023{\natexlab{a}})Fu, Zhang, Kwon, Perera, Zhu, Zhang, Li, Wang, Wang, Castelli et~al.}]{fu2023generate}
Xingyu Fu, Sheng Zhang, Gukyeong Kwon, Pramuditha Perera, Henghui Zhu, Yuhao Zhang, Alexander~Hanbo Li, William~Yang Wang, Zhiguo Wang, Vittorio Castelli, et~al. 2023{\natexlab{a}}.
\newblock Generate then select: Open-ended visual question answering guided by world knowledge.
\newblock \emph{arXiv preprint arXiv:2305.18842}.

\bibitem[{Fu et~al.(2022)Fu, Zhou, Chandratreya, Vondrick, and Roth}]{fu2022there}
Xingyu Fu, Ben Zhou, Ishaan Chandratreya, Carl Vondrick, and Dan Roth. 2022.
\newblock There’s a time and place for reasoning beyond the image.
\newblock In \emph{Proceedings of the 60th Annual Meeting of the Association for Computational Linguistics (Volume 1: Long Papers)}, pages 1138--1149.

\bibitem[{Fu et~al.(2023{\natexlab{b}})Fu, Zhou, Chen, Yatskar, and Roth}]{fu2023interpretable}
Xingyu Fu, Ben Zhou, Sihao Chen, Mark Yatskar, and Dan Roth. 2023{\natexlab{b}}.
\newblock Interpretable by design visual question answering.
\newblock \emph{arXiv preprint arXiv:2305.14882}.

\bibitem[{Goyal et~al.(2017)Goyal, Khot, Summers-Stay, Batra, and Parikh}]{goyal2017making}
Yash Goyal, Tejas Khot, Douglas Summers-Stay, Dhruv Batra, and Devi Parikh. 2017.
\newblock Making the v in vqa matter: Elevating the role of image understanding in visual question answering.
\newblock In \emph{Proceedings of the IEEE conference on computer vision and pattern recognition}, pages 6904--6913.

\bibitem[{Hu et~al.(2023{\natexlab{a}})Hu, Zhang, Zhao, and Sun}]{hu2023ciem}
Hongyu Hu, Jiyuan Zhang, Minyi Zhao, and Zhenbang Sun. 2023{\natexlab{a}}.
\newblock Ciem: Contrastive instruction evaluation method for better instruction tuning.
\newblock \emph{arXiv preprint arXiv:2309.02301}.

\bibitem[{Hu et~al.(2022)Hu, Hua, Yang, Shi, Smith, and Luo}]{hu2022promptcap}
Yushi Hu, Hang Hua, Zhengyuan Yang, Weijia Shi, Noah~A Smith, and Jiebo Luo. 2022.
\newblock Promptcap: Prompt-guided task-aware image captioning.
\newblock \emph{arXiv preprint arXiv:2211.09699}.

\bibitem[{Hu et~al.(2023{\natexlab{b}})Hu, Liu, Kasai, Wang, Ostendorf, Krishna, and Smith}]{hu2023tifa}
Yushi Hu, Benlin Liu, Jungo Kasai, Yizhong Wang, Mari Ostendorf, Ranjay Krishna, and Noah~A Smith. 2023{\natexlab{b}}.
\newblock Tifa: Accurate and interpretable text-to-image faithfulness evaluation with question answering.
\newblock In \emph{Proceedings of the IEEE/CVF International Conference on Computer Vision}, pages 20406--20417.

\bibitem[{Krishna et~al.(2017)Krishna, Zhu, Groth, Johnson, Hata, Kravitz, Chen, Kalantidis, Li, Shamma et~al.}]{krishna2017visual}
Ranjay Krishna, Yuke Zhu, Oliver Groth, Justin Johnson, Kenji Hata, Joshua Kravitz, Stephanie Chen, Yannis Kalantidis, Li-Jia Li, David~A Shamma, et~al. 2017.
\newblock Visual genome: Connecting language and vision using crowdsourced dense image annotations.
\newblock \emph{International journal of computer vision}, 123:32--73.

\bibitem[{Li et~al.(2023)Li, Du, Zhou, Wang, Zhao, and Wen}]{li2023evaluating}
Yifan Li, Yifan Du, Kun Zhou, Jinpeng Wang, Wayne~Xin Zhao, and Ji-Rong Wen. 2023.
\newblock Evaluating object hallucination in large vision-language models.
\newblock \emph{arXiv preprint arXiv:2305.10355}.

\bibitem[{Lin et~al.(2014)Lin, Maire, Belongie, Hays, Perona, Ramanan, Doll{\'a}r, and Zitnick}]{lin2014microsoft}
Tsung-Yi Lin, Michael Maire, Serge Belongie, James Hays, Pietro Perona, Deva Ramanan, Piotr Doll{\'a}r, and C~Lawrence Zitnick. 2014.
\newblock Microsoft coco: Common objects in context.
\newblock In \emph{Computer Vision--ECCV 2014: 13th European Conference, Zurich, Switzerland, September 6-12, 2014, Proceedings, Part V 13}, pages 740--755. Springer.

\bibitem[{Liu et~al.(2023{\natexlab{a}})Liu, Guan, Li, Chen, Yacoob, Manocha, and Zhou}]{liu2023hallusionbench}
Fuxiao Liu, Tianrui Guan, Zongxia Li, Lichang Chen, Yaser Yacoob, Dinesh Manocha, and Tianyi Zhou. 2023{\natexlab{a}}.
\newblock Hallusionbench: You see what you think? or you think what you see? an image-context reasoning benchmark challenging for gpt-4v (ision), llava-1.5, and other multi-modality models.
\newblock \emph{arXiv preprint arXiv:2310.14566}.

\bibitem[{Liu et~al.(2023{\natexlab{b}})Liu, Lin, Li, Wang, Yacoob, and Wang}]{liu2023aligning}
Fuxiao Liu, Kevin Lin, Linjie Li, Jianfeng Wang, Yaser Yacoob, and Lijuan Wang. 2023{\natexlab{b}}.
\newblock Aligning large multi-modal model with robust instruction tuning.
\newblock \emph{arXiv preprint arXiv:2306.14565}.

\bibitem[{Liu et~al.(2023{\natexlab{c}})Liu, Li, Wu, and Lee}]{liu2023llava}
Haotian Liu, Chunyuan Li, Qingyang Wu, and Yong~Jae Lee. 2023{\natexlab{c}}.
\newblock Visual instruction tuning.

\bibitem[{Liu et~al.(2024)Liu, Li, Wu, and Lee}]{liu2024visual}
Haotian Liu, Chunyuan Li, Qingyang Wu, and Yong~Jae Lee. 2024.
\newblock Visual instruction tuning.
\newblock \emph{Advances in neural information processing systems}, 36.

\bibitem[{Lovenia et~al.(2023)Lovenia, Dai, Cahyawijaya, Ji, and Fung}]{lovenia2023negative}
Holy Lovenia, Wenliang Dai, Samuel Cahyawijaya, Ziwei Ji, and Pascale Fung. 2023.
\newblock Negative object presence evaluation (nope) to measure object hallucination in vision-language models.
\newblock \emph{arXiv preprint arXiv:2310.05338}.

\bibitem[{OpenAI(2023{\natexlab{a}})}]{openai2023gpt}
R~OpenAI. 2023{\natexlab{a}}.
\newblock Gpt-4 technical report. arxiv 2303.08774.
\newblock \emph{View in Article}, 2(5).

\bibitem[{OpenAI(2023{\natexlab{b}})}]{openai2023gptsys}
R~OpenAI. 2023{\natexlab{b}}.
\newblock Gpt-4v (ision) system card.
\newblock \emph{Citekey: gptvision}.

\bibitem[{Pryzant et~al.(2023)Pryzant, Iter, Li, Lee, Zhu, and Zeng}]{pryzant2023automatic}
Reid Pryzant, Dan Iter, Jerry Li, Yin~Tat Lee, Chenguang Zhu, and Michael Zeng. 2023.
\newblock Automatic prompt optimization with" gradient descent" and beam search.
\newblock \emph{arXiv preprint arXiv:2305.03495}.

\bibitem[{Radford et~al.(2021)Radford, Kim, Hallacy, Ramesh, Goh, Agarwal, Sastry, Askell, Mishkin, Clark et~al.}]{radford2021learning}
Alec Radford, Jong~Wook Kim, Chris Hallacy, Aditya Ramesh, Gabriel Goh, Sandhini Agarwal, Girish Sastry, Amanda Askell, Pamela Mishkin, Jack Clark, et~al. 2021.
\newblock Learning transferable visual models from natural language supervision.
\newblock In \emph{International conference on machine learning}, pages 8748--8763. PMLR.

\bibitem[{Reid et~al.(2024)Reid, Savinov, Teplyashin, Lepikhin, Lillicrap, Alayrac, Soricut, Lazaridou, Firat, Schrittwieser et~al.}]{reid2024gemini}
Machel Reid, Nikolay Savinov, Denis Teplyashin, Dmitry Lepikhin, Timothy Lillicrap, Jean-baptiste Alayrac, Radu Soricut, Angeliki Lazaridou, Orhan Firat, Julian Schrittwieser, et~al. 2024.
\newblock Gemini 1.5: Unlocking multimodal understanding across millions of tokens of context.
\newblock \emph{arXiv preprint arXiv:2403.05530}.

\bibitem[{Russakovsky et~al.(2015)Russakovsky, Deng, Su, Krause, Satheesh, Ma, Huang, Karpathy, Khosla, Bernstein et~al.}]{russakovsky2015imagenet}
Olga Russakovsky, Jia Deng, Hao Su, Jonathan Krause, Sanjeev Satheesh, Sean Ma, Zhiheng Huang, Andrej Karpathy, Aditya Khosla, Michael Bernstein, et~al. 2015.
\newblock Imagenet large scale visual recognition challenge.
\newblock \emph{International journal of computer vision}, 115:211--252.

\bibitem[{Tong et~al.(2024)Tong, Liu, Zhai, Ma, LeCun, and Xie}]{tong2024eyes}
Shengbang Tong, Zhuang Liu, Yuexiang Zhai, Yi~Ma, Yann LeCun, and Saining Xie. 2024.
\newblock Eyes wide shut? exploring the visual shortcomings of multimodal llms.
\newblock \emph{arXiv preprint arXiv:2401.06209}.

\bibitem[{Touvron et~al.(2023)Touvron, Martin, Stone, Albert, Almahairi, Babaei, Bashlykov, Batra, Bhargava, Bhosale et~al.}]{touvron2023llama}
Hugo Touvron, Louis Martin, Kevin Stone, Peter Albert, Amjad Almahairi, Yasmine Babaei, Nikolay Bashlykov, Soumya Batra, Prajjwal Bhargava, Shruti Bhosale, et~al. 2023.
\newblock Llama 2: Open foundation and fine-tuned chat models.
\newblock \emph{arXiv preprint arXiv:2307.09288}.

\bibitem[{Wei et~al.(2022)Wei, Wang, Schuurmans, Bosma, Xia, Chi, Le, Zhou et~al.}]{wei2022chain}
Jason Wei, Xuezhi Wang, Dale Schuurmans, Maarten Bosma, Fei Xia, Ed~Chi, Quoc~V Le, Denny Zhou, et~al. 2022.
\newblock Chain-of-thought prompting elicits reasoning in large language models.
\newblock \emph{Advances in neural information processing systems}, 35:24824--24837.

\bibitem[{Yang et~al.(2023)Yang, Li, Lin, Wang, Lin, Liu, and Wang}]{yang2023dawn}
Zhengyuan Yang, Linjie Li, Kevin Lin, Jianfeng Wang, Chung-Ching Lin, Zicheng Liu, and Lijuan Wang. 2023.
\newblock The dawn of lmms: Preliminary explorations with gpt-4v (ision).
\newblock \emph{arXiv preprint arXiv:2309.17421}, 9(1):1.

\bibitem[{Yuan et~al.(2024)Yuan, Jiao, Wang, Huang, Xu, Liang, He, and Tu}]{yuan2024refuse}
Youliang Yuan, Wenxiang Jiao, Wenxuan Wang, Jen-tse Huang, Jiahao Xu, Tian Liang, Pinjia He, and Zhaopeng Tu. 2024.
\newblock Refuse whenever you feel unsafe: Improving safety in llms via decoupled refusal training.
\newblock \emph{arXiv preprint arXiv:2407.09121}.

\bibitem[{Zhang et~al.(2024)Zhang, Jiang, Zhang, Lin, Guo, Qiu, Zhou, Lu, Chang, Gao et~al.}]{zhang2024mathverse}
Renrui Zhang, Dongzhi Jiang, Yichi Zhang, Haokun Lin, Ziyu Guo, Pengshuo Qiu, Aojun Zhou, Pan Lu, Kai-Wei Chang, Peng Gao, et~al. 2024.
\newblock Mathverse: Does your multi-modal llm truly see the diagrams in visual math problems?
\newblock \emph{arXiv preprint arXiv:2403.14624}.

\end{thebibliography}

% \clearpage
\appendix
\section{Experiement and Evaluation Details}
We provide more details of our evaluations and experiments in this section.
\subsection{Why Description is a ``More Difficult Task''?}
\label{why hard}
We categorize binary (Yes/No) and multiple-choice questions as ``simple questions'' from the perspective of the probability of random guessing and the size of the solution space. For a Yes/No question, there is a 50\% chance of guessing correctly, and for a multiple-choice question with three options, there is a 33\% chance of guessing correctly. Conversely, an open-ended question such as ``Describe the man's outfit of his upper body'' requires the model to generate a highly specific and correct response from an infinite combination of characters, such as ``He is wearing a sleeveless tank top'' or ``a vest,'' to be considered “correct”. In this case, the probability of a correctly random guess is nearly zero. Therefore, intuitively, we believe that accurately describing an object is more challenging than selecting the correct answer from a limited set of options. Since we do not focus on ``reasoning difficulty,'' all our questions are specifically designed to ensure that the ``reasoning difficulty'' of Yes/No questions is comparable to that of descriptive questions. For example, in Figure 1(a), the descriptive task ``Please describe the cap that the man is wearing'' does not cover more information than the Yes/No question ``Is the man wearing a beige cap?'' and they are asking about the same thing. In a word, the term ``difficult task'' here refers solely to the type of question, not the specific question content.

\subsection{Conversed \& Irrelevant Question for Ablation Studies}
\label{sec:example of conv}
\begin{figure}[htbp]
  \centering
  \includegraphics[width=0.5\textwidth]{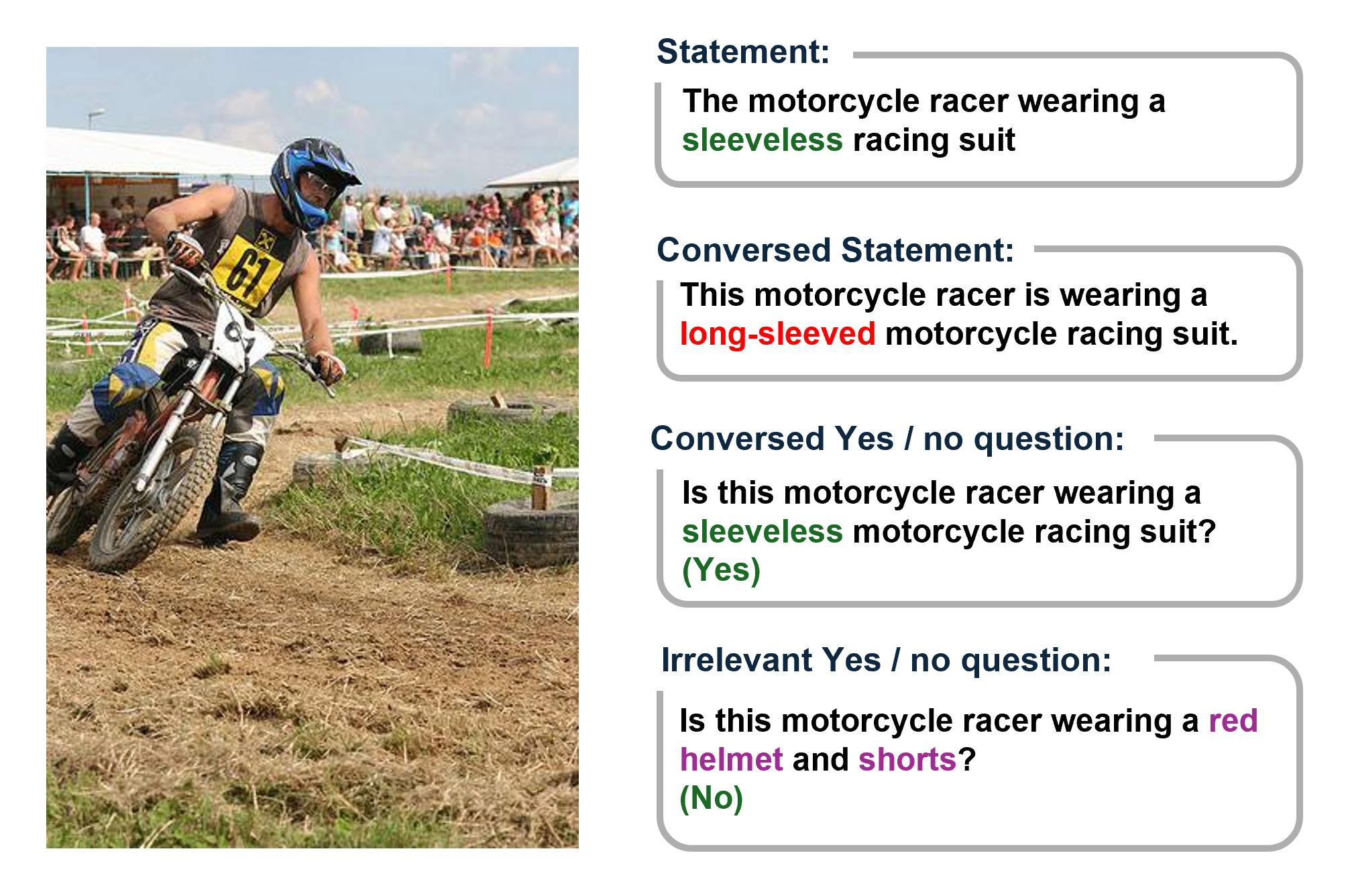}
  \caption{The statement is a brief statement about the ``question subject'' in the image. The conversed statement contradicts the ``question subject''. The irrelevant question is a  Yes/No question unrelated to the image content, and the conversed Yes/No question is derived from the correct statement. They are used to ensure that the model does not thoughtlessly respond with ``yes.''}
  \label{fig: example of conversed questions}
\end{figure}

We use GPT-4-turbo to generate a conversed Yes/No question and an irrelevant Yes/No question for each image, based on the statements. As shown in Figure \ref{fig: example of conversed questions}, the irrelevant questions are about something unrelated to the images, and their ground truth answers are all ``No.'' If a model always fails in the irrelevant questions, we believe it tends to give ``yes'' responses without hesitation. All the ground truth answers to the conversed Yes/No questions are ``yes''. They test if MLLMs can correctly recognize the features which are indeed in this image. 

\begin{figure*}[htbp]
  \centering
    \includegraphics[width=0.95\linewidth]{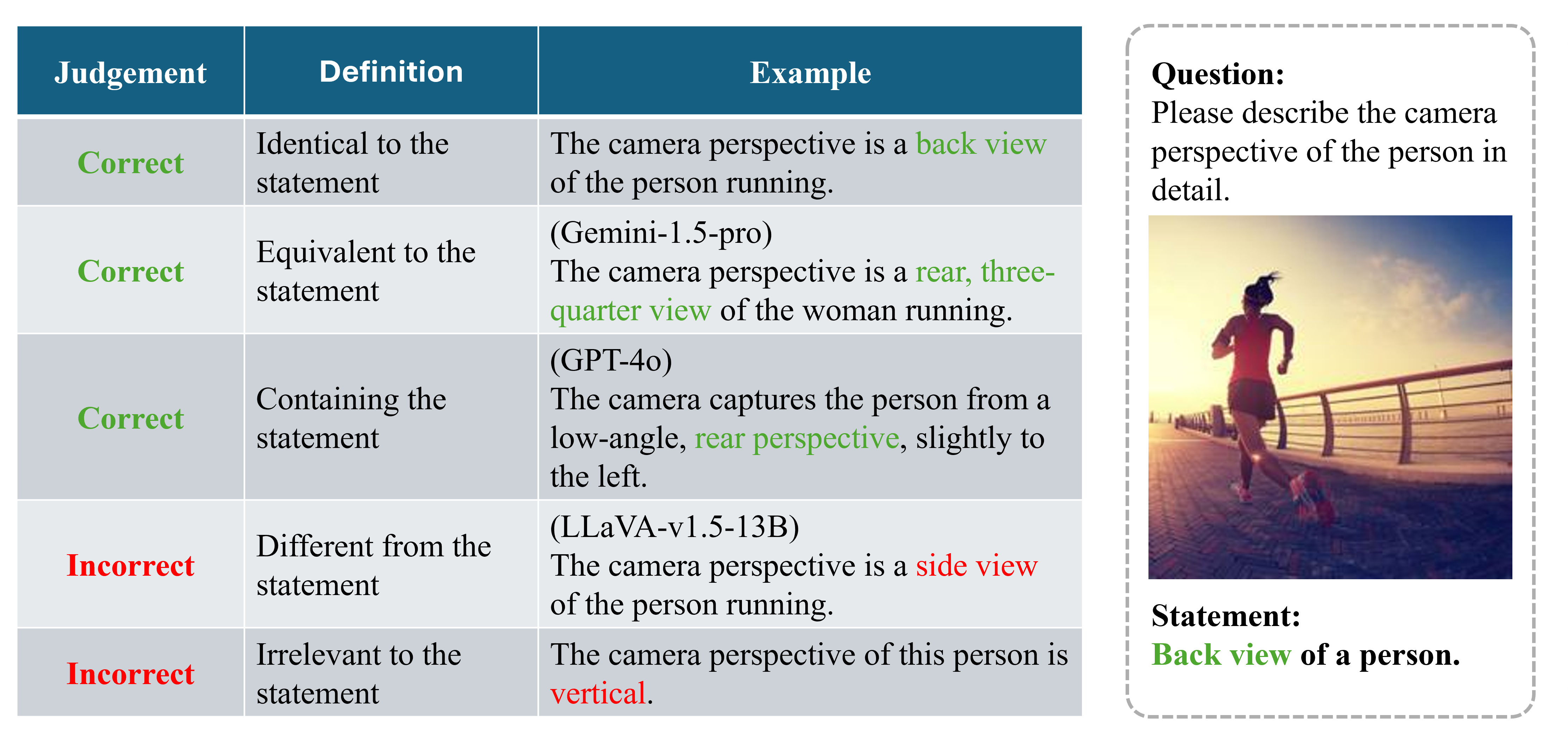}
    \caption{Example of the evaluation criterion of the description.}
    \label{fig:criterion}
\end{figure*}

\subsection{Swapping Options}
We use the swapping option to ensure that the result in multiple choice is not influenced by option bias of ``A'', ``B'' and ``C''. Empirically we find that LLaVA-1.5-7B obtains 91.21\% accuracy when the correct answer is ``A'', but 3.2\% accuracy when the correct answer is ``B'' on LazyBench. Other MLLMs tend to have even performance while the order of the options shifts.

\subsection{Evaluation Criterion of the Description}
\label{Evaluation Criterion}
We employ a binary classification method to score the descriptions provided by the model as either correct or incorrect. As shown in Figure \ref{fig:criterion}, when the MLLM gives a description identical to the statement, we judge it as correct (e.g., if the MLLM's description: ``This is a back view of a person'' and the statement: ``back view of a person''). If the MLLM provides a description different from the statement, but we find it equivalent to the statement or containing the statement when considering the image, we also judge it as correct (e.g., GPT-4o's description: ``The camera captures the person from a low-angle, rear perspective, slightly to the left,'' which we consider a more detailed description based on the image). In all other cases, if the MLLM's description differs from the statement and is neither equivalent nor contains the same information, we judge it as incorrect (e.g., LLaVA-1.5-13B's description: ``The camera perspective is a side view of the person running''). Additionally, if the model's description is irrelevant to our question or refuses to answer the relevant question, we also consider it an incorrect description. 

\subsection{Prompts of Doby}
\begin{figure*}[htbp]
  \centering
  \includegraphics[width=0.95\textwidth]{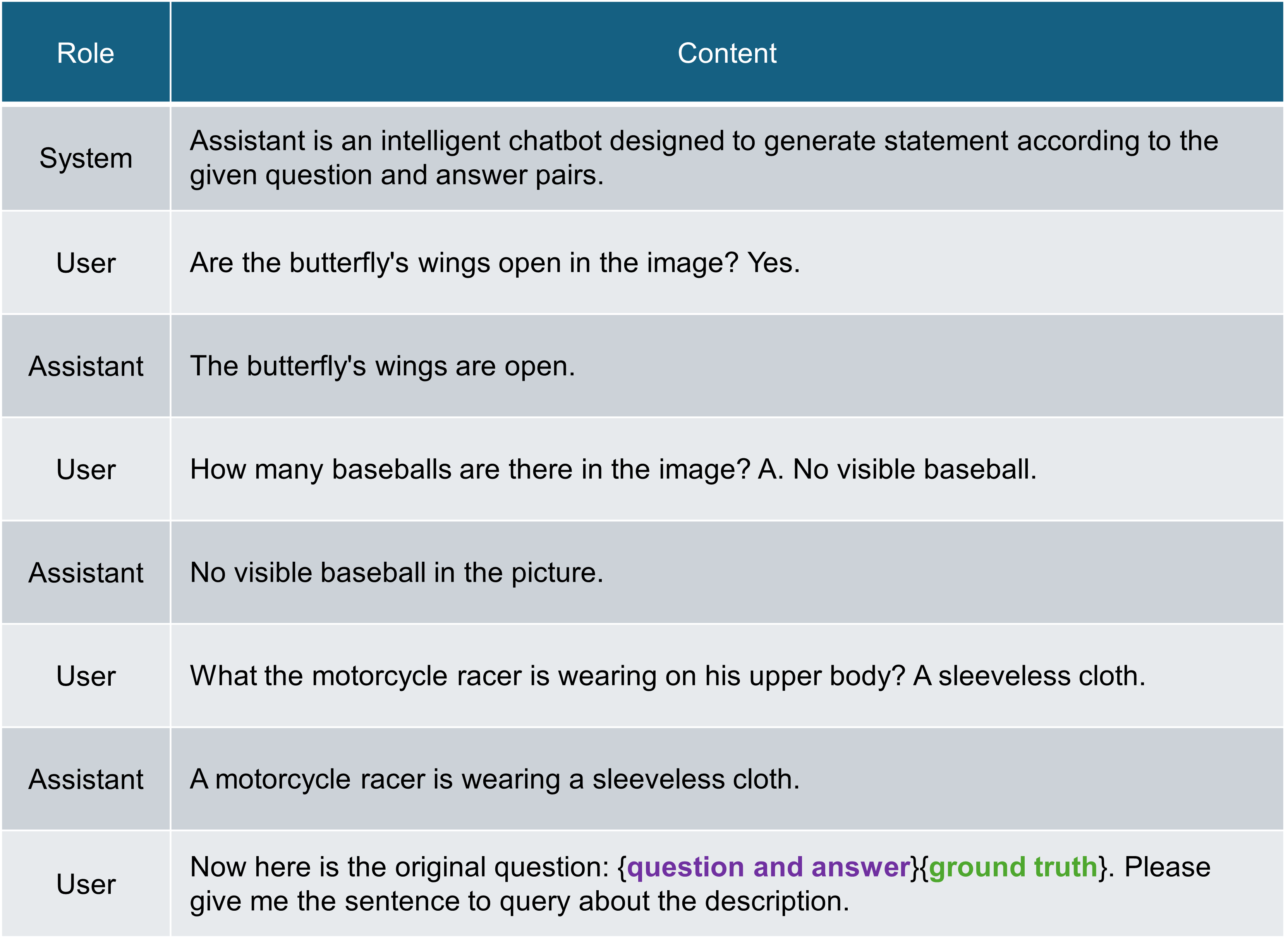}
  \caption{Example of the few-shot prompt to generate the statement.}
  \label{fig: doby_1}
\end{figure*}

\begin{figure*}[htbp]
  \centering
  \includegraphics[width=0.95\textwidth]{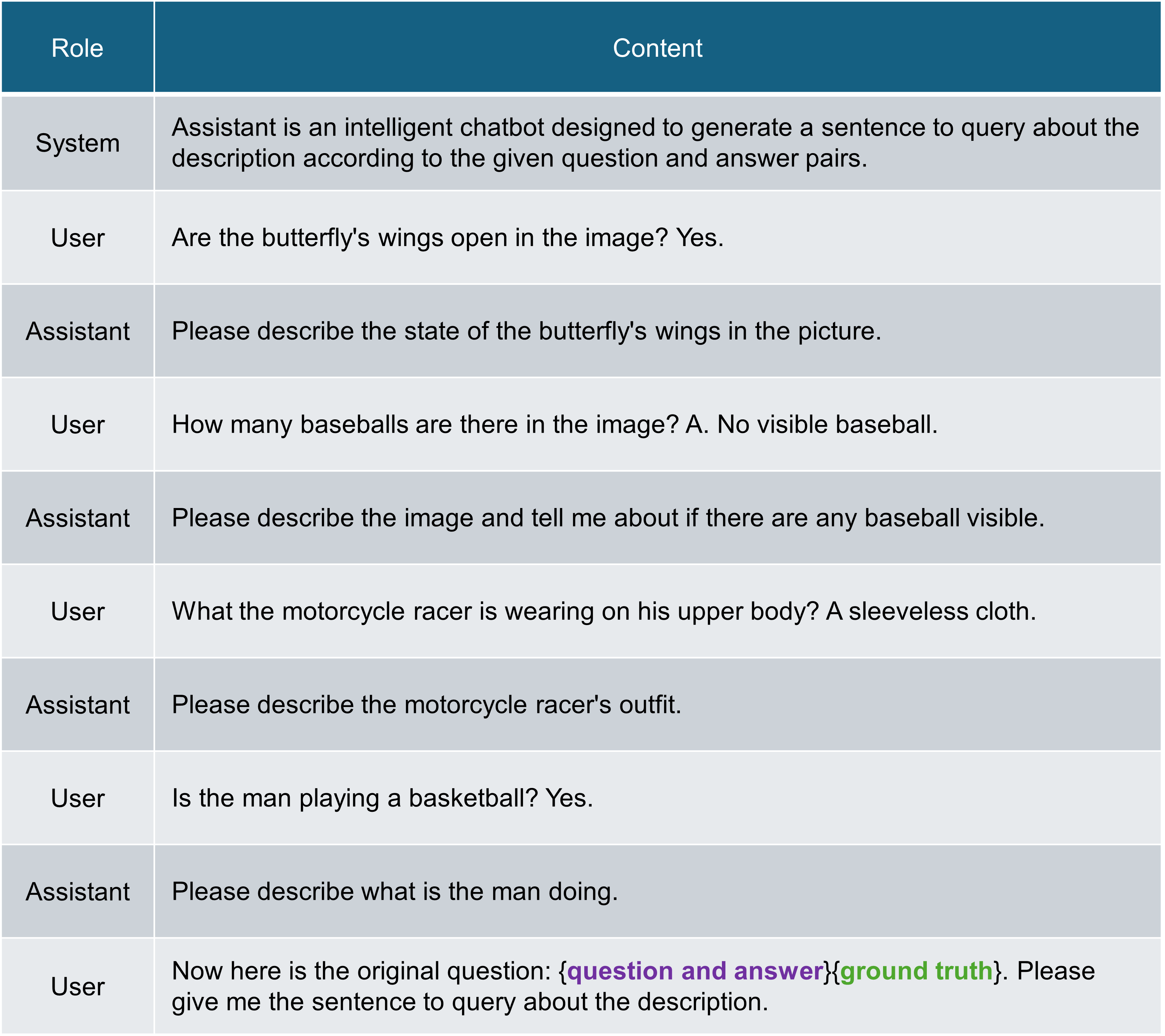}
  \caption{Example of the few-shot prompt to generate the description question.}
  \label{fig: doby_2}
\end{figure*}

In Doby, we first ask GPT-4-turbo to generate the statement by Yes/No question and answer pairs, here we use a few shot prompt strategy (Figure \ref{fig: doby_1}). Then we use GPT-4-turbo to generate the description request (Figure \ref{fig: doby_2}). After asking the MLLMs to answer the description request, Doby compared the statements and the MLLMs' descriptions (by using GPT-4-turbo, with the same criterion in Figure \ref{fig:criterion}) to check if MLLMs can correctly describe the subject in the image.

% \clearpage

\section{More Results of LazyBench and Doby}
\label{sec:hallusionbench}
\begin{figure*}[htbp]
  \centering
    \includegraphics[width=0.95\linewidth]{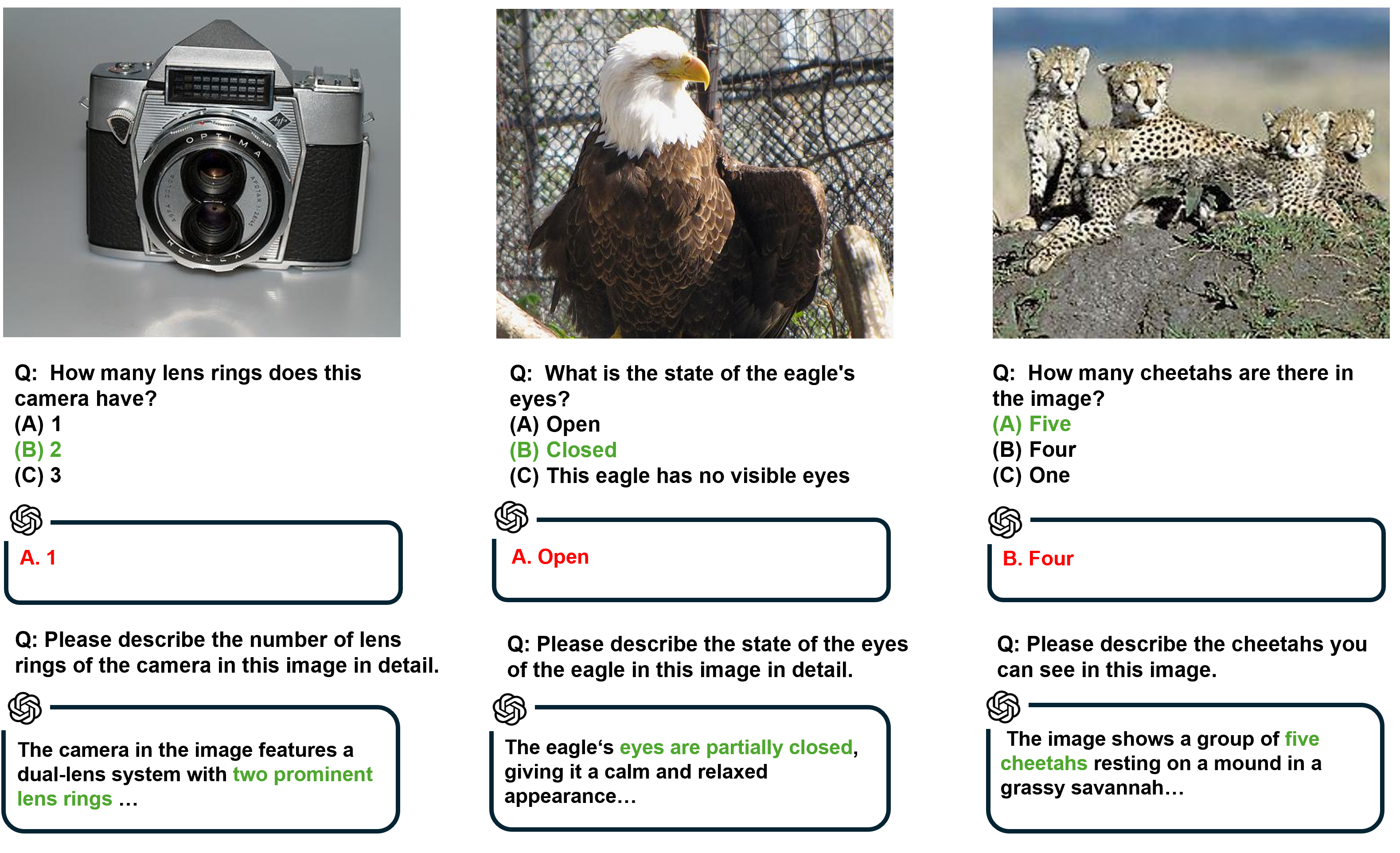}
    \caption{More examples of GPT-4V's laziness in LazyBench (Multiple choice).}
    \label{fig:example of multi}
    
\end{figure*}

\begin{figure*}[htbp]
  \centering
    \includegraphics[width=0.95\linewidth]{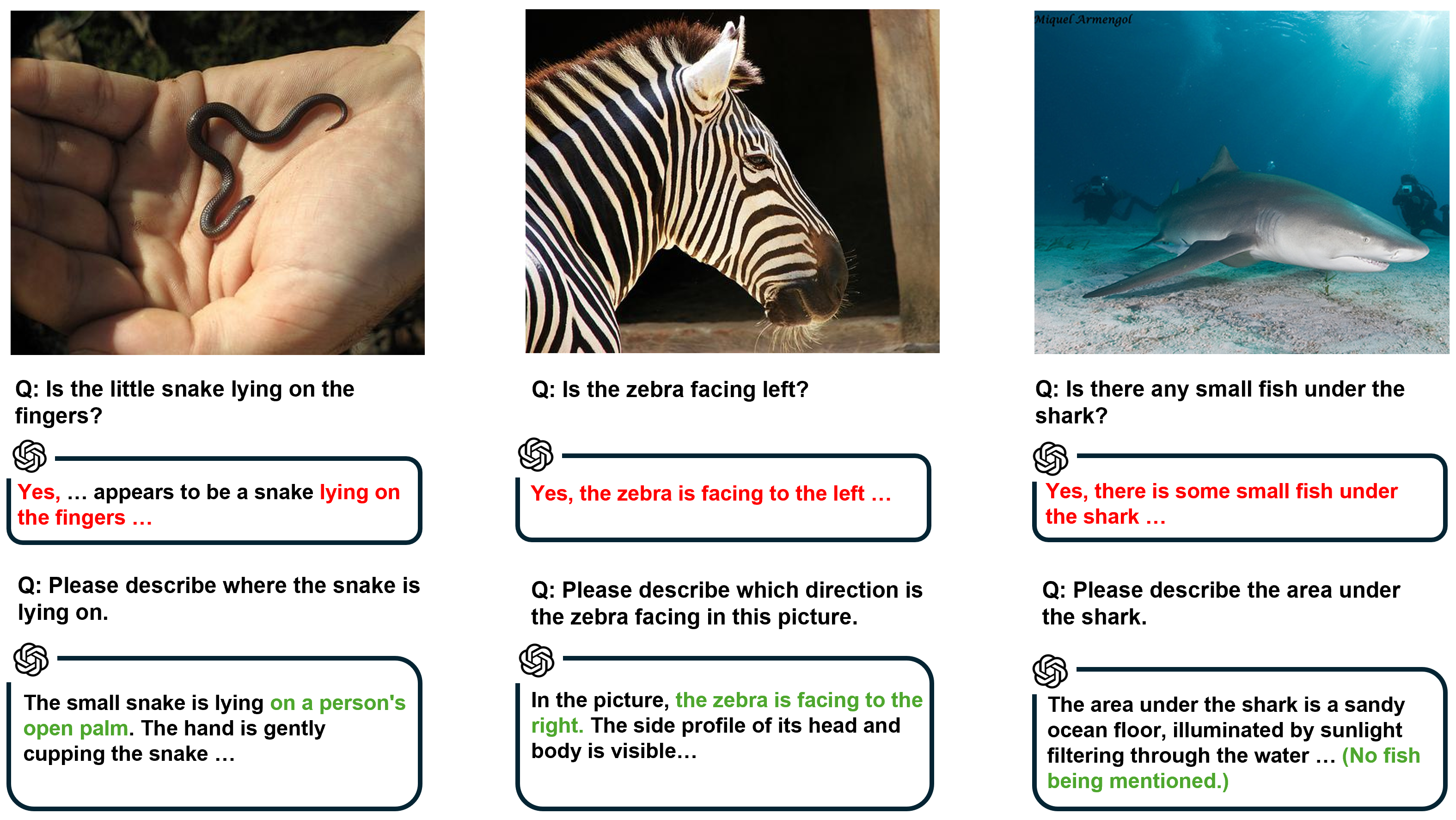}
    \caption{More examples of GPT-4V's laziness in LazyBench (Yes/No question).}
    \label{fig:example of yes}
\end{figure*}

Here we display more results of MLLMs' performance on LazyBench (Figure \ref{fig:example of multi}, \ref{fig:example of yes}) and the findings given by Doby.
% This Section contains the failure cases in Hallusionbench, we manually checked the processed result of Doby and find there are 


\begin{figure*}[htbp]
  \centering
    \includegraphics[width=0.95\linewidth]{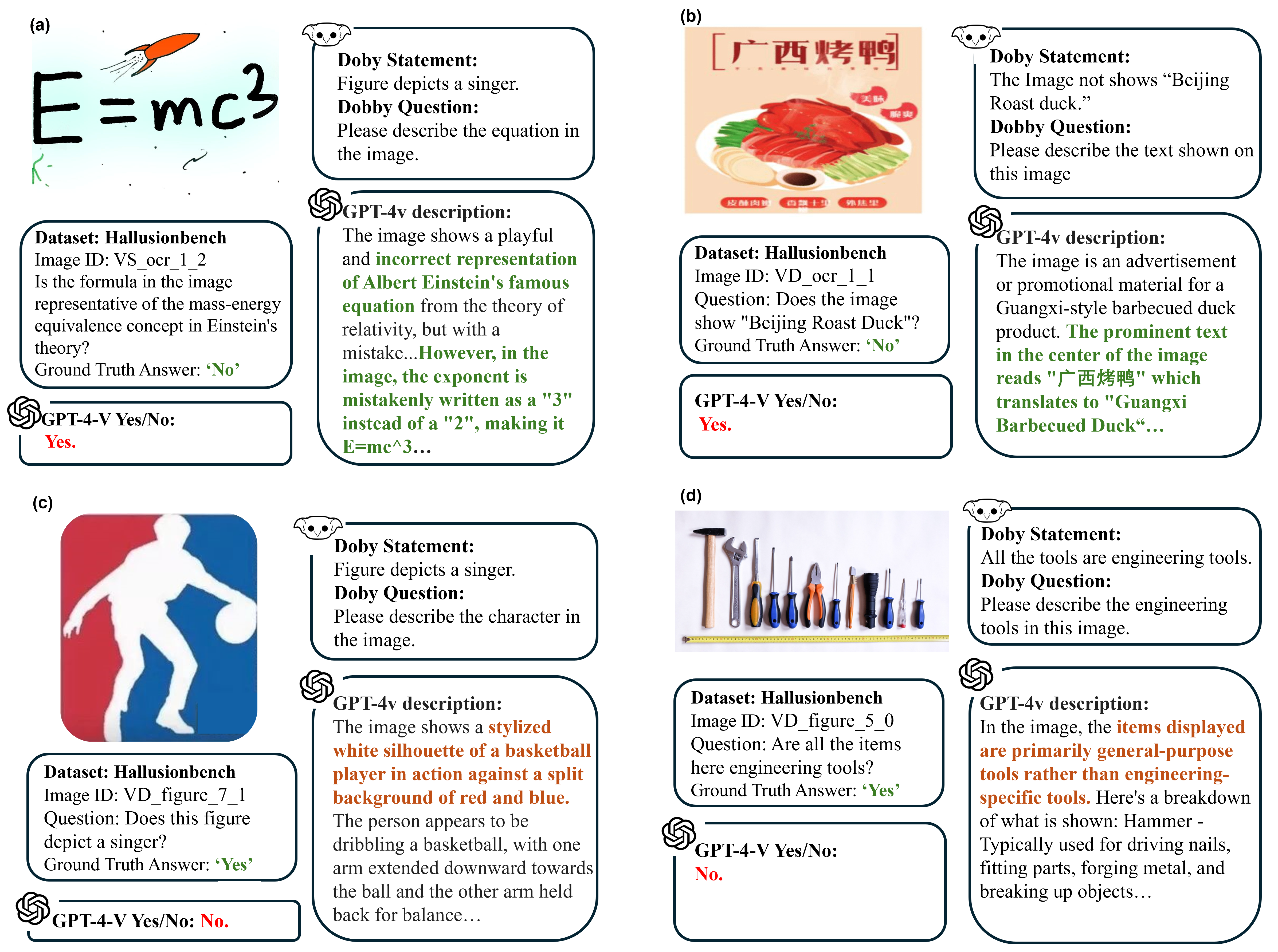}
    \caption{The examples of GPT-4V's failure cases in Hallusionbench \cite{liu2023hallusionbench}. (a)(b) GPT-4V is being lazy when answering the original questions. (c) The original visual information is ambiguous. (d) The ambiguous definition of the ``engineer tool'' in the original question causes the failure.}
    \label{fig:fig 5}
\end{figure*}

Using Hallusionbench \cite{liu2023hallusionbench} as an example: In their work, the case in Figure \ref{fig:fig 5}(c) will be a case of visual illusion (if the model correctly identifies the previous question about the original NBA logo as a basketball player.) However, our Doby shows that in these cases, MLLMs make mistakes for other reasons (i.e., in this case, lack of related information. MLLMs can correctly describe every element in the image and human beings who do not know the character cannot tell this person in the logo is a singer either.) We do not expect MLLMs to know everything knowledge so we cannot sorely define the mistake as "Hallucination". 

\end{document}